\documentclass[sigconf]{acmart}
\AtBeginDocument{%
  }

\copyrightyear{2025}
\acmYear{2025}
\setcopyright{cc}
\setcctype{by}
\acmConference[KDD '25]{Proceedings of the 31st ACM SIGKDD Conference on Knowledge Discovery and Data Mining V.2}{August 3--7, 2025}{Toronto, ON, Canada}
\acmBooktitle{Proceedings of the 31st ACM SIGKDD Conference on Knowledge Discovery and Data Mining V.2 (KDD '25), August 3--7, 2025, Toronto, ON, Canada}
\acmDOI{10.1145/3711896.3736849}
\acmISBN{979-8-4007-1454-2/2025/08}

\usepackage{xcolor}
\definecolor{darkred}{RGB}{160,0,0}
\definecolor{darkgreen}{RGB}{0,130,51}

\usepackage{multirow}

\usepackage{amssymb}
\usepackage{graphicx}

\usepackage{array}

\usepackage{natbib}
\usepackage{xspace}

\usepackage{listings}

\usepackage{algorithm}
\usepackage{algpseudocode}

\usepackage{makecell}
\usepackage{hyperref}
\usepackage{cleveref}

\newcommand{\dataset}{\textsc{BlendQA}\xspace}
\newcommand{\model}{\textsc{AtomR}\xspace}
\newcolumntype{C}[1]{>{\centering\arraybackslash}p{#1}} 

\newcommand{\search}{\texttt{Search}\xspace}
\newcommand{\relate}{\texttt{Relate}\xspace}
\newcommand{\filter}{\texttt{Filter}\xspace}

\begin{document}

\title{\model: Atomic Operator-Empowered Large Language Models for Heterogeneous Knowledge Reasoning}

\author{Amy Xin}
\authornote{Equal contribution.}
\email{xin-x23@mails.tsinghua.edu.cn}
\orcid{0009-0001-2404-0475}
\affiliation{%
  \institution{Tsinghua University}
  \city{Beijing}
  \country{China}
}

\author{Jinxin Liu}

\email{liujinxi20@mails.tsinghua.edu.cn}
\authornotemark[1]
\affiliation{%
  \institution{Tsinghua University}
  \city{Beijing}
  \country{China}
}

\author{Zijun Yao}
\email{yaozj20@mails.tsinghua.edu.cn}
\affiliation{%
  \institution{Tsinghua University}
  \city{Beijing}
  \country{China}
}

\author{Zhicheng Lee}
\email{lizhiche23@mails.tsinghua.edu.cn}
\affiliation{%
  \institution{Tsinghua University}
  \city{Beijing}
  \country{China}
}

\author{Shulin Cao}
\email{shulincao97@163.com}
\affiliation{%
  \institution{Zhipu AI}
  \city{Beijing}
  \country{China}
}

\author{Lei Hou}
\authornote{Corresponding Authors.}
\email{houlei@tsinghua.edu.cn}
\affiliation{%
  \institution{Tsinghua University}
  \city{Beijing}
  \country{China}}

\author{Juanzi Li}
\authornotemark[2]
\email{lijuanzi@tsinghua.edu.cn}
\affiliation{%
  \institution{Tsinghua University}
  \city{Beijing}
  \country{China}}

\renewcommand{\shortauthors}{Amy Xin et al.}


\begin{abstract}

Despite the outstanding capabilities of large language models (LLMs), knowledge-intensive reasoning still remains a challenging task due to LLMs' limitations in compositional reasoning and the hallucination problem. 
A prevalent solution is to employ chain-of-thought (CoT) with retrieval-augmented generation (RAG), which first formulates a reasoning plan by decomposing complex questions into simpler sub-questions, and then applies iterative RAG at each sub-question.
However, prior works exhibit two crucial problems: inadequate reasoning planning and poor incorporation of heterogeneous knowledge. 
In this paper, we introduce \textbf{\model}, a framework for LLMs to conduct accurate heterogeneous knowledge reasoning
at the \textit{atomic} level.
Inspired by how knowledge graph query languages model compositional reasoning through combining predefined operations,
we propose three \textit{atomic knowledge operators}, a unified set of operators for LLMs to retrieve and manipulate knowledge from heterogeneous sources.
First, in the reasoning planning stage, \model decomposes a complex question into a reasoning tree where each leaf node corresponds to an atomic knowledge operator, achieving question decomposition that is highly fine-grained and orthogonal.
Subsequently, in the reasoning execution stage, \model 
executes each atomic knowledge operator, which flexibly selects, retrieves, and operates atomic level knowledge from heterogeneous sources.
We also introduce \textbf{\dataset}, a challenging benchmark specially tailored for heterogeneous knowledge reasoning.
Experiments on three single-source and two multi-source datasets show that \model outperforms state-of-the-art baselines by a large margin, with absolute F1 score improvements of $9.4\%$ on 2WikiMultihop and $9.5\%$ on \dataset.
We release our code and data\footnote{\href{https://github.com/THU-KEG/AtomR.git}{https://github.com/THU-KEG/AtomR.git}}.
\end{abstract}

\ccsdesc[500]{Information systems~Question answering}
\ccsdesc[300]{Computing methodologies~Natural language generation}

\keywords{Retrieval-Augmented Generation, Large Language Models, Knowledge-Intensive Reasoning, Multi-hop QA}



\maketitle

\newcommand\kddavailabilityurl{https://doi.org/10.5281/zenodo.15528472}

\ifdefempty{\kddavailabilityurl}{}{
\begingroup\small\noindent\raggedright\textbf{KDD Availability Link:}\\
The source code and data of this paper has been made publicly available at \url{\kddavailabilityurl}.
\endgroup
}

\section{Introduction}

Knowledge-intensive reasoning is a challenging task that requires the ability to perform compositional reasoning over vast volumes of knowledge, demanding skills such as multi-hop inference, comparison and calculation~\citep{DBLP:journals/corr/abs-2202-08772, DBLP:conf/www/GuKVSLY021, DBLP:conf/acl/CaoSPNX0LHZ22, musique}.
Although recent advancements of Large Language Models (LLMs)~\citep{DBLP:journals/corr/abs-2303-08774} have enabled them to excel in various natural language processing tasks\citep{DBLP:journals/corr/abs-2303-18223, DBLP:journals/corr/abs-2302-04023}, 
it is still arduous for LLMs to perform knowledge-intensive reasoning due to their weaknesses in implicit compositional reasoning~\citep{DBLP:conf/emnlp/PressZMSSL23, dziri2024faith}
and the hallucination problem~\citep{DBLP:journals/corr/abs-2311-05232, DBLP:journals/corr/abs-2302-04023}.
To address these issues, recent works~\citep{DBLP:conf/acl/TrivediBKS23, DBLP:conf/emnlp/PressZMSSL23, DBLP:conf/www/XuPSCC24, DBLP:conf/emnlp/CaoZSL0THL23} propose to leverage chain-of-thought (CoT)~\citep{DBLP:conf/nips/Wei0SBIXCLZ22} reasoning with retrieval augmented generation (RAG):
specifically, to first perform question decomposition on complex questions into simpler sub-questions, thereby formulating an explicit reasoning plan, and then retrieve external knowledge for sub-question answering, thereby alleviating LLM hallucination.

However, there are still three main challenges. 
$\mathbf{\mathcal{C}1}$. \textbf{Sub-optimal reasoning planning due to inadequate question decomposition}.
Existing works have explored question decomposition into different structures, such as chain~\citep{DBLP:conf/acl/TrivediBKS23, DBLP:conf/emnlp/PressZMSSL23, DBLP:conf/emnlp/ShaoGSHDC23, DBLP:conf/www/XuPSCC24} and tree structures~\citep{DBLP:conf/nips/YaoYZS00N23, DBLP:conf/emnlp/CaoZSL0THL23}. 
However, despite constraining the decomposition structure, previous works leverage LLMs to generate sub-questions \textit{freely}, imposing no supervision over the atomicity or semantic function of sub-questions.
This results in the generation of coarse-grained sub-questions as well as sub-questions with inaccurate functionalities, leading to sub-optimal reasoning paths and mistakes.
$\mathbf{\mathcal{C}2}$. \textbf{The limited support for multiple heterogeneous knowledge sources}.
Heterogeneous knowledge sources such as online web pages, local text corpora, and structured knowledge bases contain rich knowledge that complement each other. 
For example, to resolve the complex question \textit{``How many studio albums has Shakira released between 2000 and 2010?''}, an effective solution is to first determine who \textit{``Shakira''} is using a web search engine, then retrieve the albums released by \textit{``Shakira''} from a structured knowledge base.
However, most existing frameworks only retrieve from a fixed knowledge source~\citep{DBLP:conf/acl/TrivediBKS23, DBLP:conf/www/XuPSCC24}. 
A few \citep{DBLP:conf/iclr/LiZCDJPB24} support multiple knowledge sources, 
but fall short in cross-source retrieval at the step level, as well as effective knowledge manipulation.
$\mathbf{\mathcal{C}3}$. \textbf{The absence of high quality datasets built on heterogeneous knowledge sources}. 
Existing benchmarks that are built on multiple knowledge sources either have a narrow knowledge scope, typically only encompassing encyclopedic knowledge from Wikipedia and Wikidata~\citep{DBLP:conf/naacl/ZhaoLNWYJZY24, DBLP:conf/acl/ZhangSGXLL24, DBLP:conf/acl/0002LLCYLH23}, or lack instance-level design, allowing individual questions to be answered using a single knowledge source without requiring cross-knowledge source querying and aggregation~\citep{DBLP:conf/www/ChristmannRW24, crag}.

In this paper, we propose \textbf{\model}, an \textbf{Atom}ic operator-empower-ed LLM \textbf{R}easoning framework for heterogeneous knowledge sources.
With the observation that explicit reasoning processes can enhance LLM reasoning, we take one more step to explore the use of \textit{explicit operators} to model 
knowledge-intensive reasoning.
Specifically, we take inspiration from how knowledge graphs (KGs) explicitly model the compositionality of knowledge
with atomic components, 
and summarize three fundamental \textit{atomic knowledge operators} for LLMs based on the grammar of existing KG query languages~\citep{DBLP:conf/acl/CaoSPNX0LHZ22, DBLP:conf/acl/YeYHZX22, DBLP:conf/www/GuKVSLY021}.
We find that all predefined operations
of 
KG query languages can be induced into a combination of three basic atomic operators: \textbf{\search}, \textbf{\relate}, and \textbf{\filter}.
These atomic operators possess the properties of indivisibility and orthogonality, meaning that each corresponds to a distinct knowledge operation without any functional overlap.
By composing these atomic operators, we can effectively model complex procedures of knowledge-intensive reasoning.
We uniquely implement each atomic knowledge operator based on its functionality, while uniformly applying them across three heterogeneous knowledge sources: local text corpora (Text), online web pages (Web), and structured knowledge graphs (KG).
During reasoning, LLMs dynamically manipulate atomic level knowledge from heterogeneous sources via this unified set of atomic operators.

\begin{figure}[t]
  \noindent
  \centering
  \includegraphics[width=\linewidth]{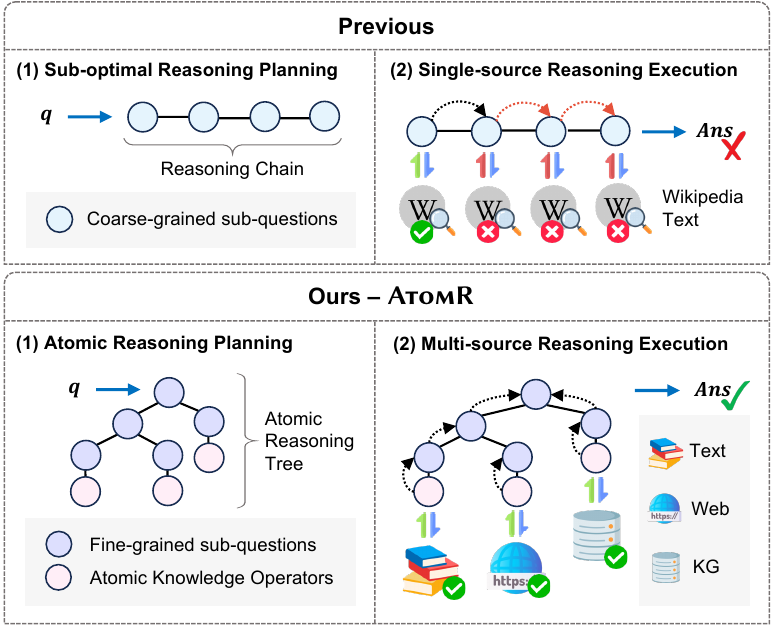}
  \caption{\label{fig:intro_figure}Comparison of \model to previous methods. 
  \model achieves more fine-grained question decomposition and manipulates knowledge from multiple heterogeneous sources.
  }
\end{figure}

As shown in Figure~\ref{fig:intro_figure}, \model effectively addresses challenges $\mathbf{\mathcal{C}1}$ and $\mathbf{\mathcal{C}2}$.
The pipeline of \model consists of two main stages: \textbf{(1) Atomic Reasoning Planning}, and \textbf{(2) Atomic Reasoning Execution}.
First, in the \textbf{Atomic Reasoning Planning} stage, \model decomposes the input question into a hierarchical \textit{Atomic Reasoning Tree} (ART), where each leaf node 
corresponds to one of the three predefined atomic knowledge operators.
This forces the LLM to continue question decomposition until an \textit{atomic} granularity is reached, achieving highly fine-grained reasoning planning with operators of distinct and orthogonal functions, effectively minimizing the impact of challenge $\mathbf{\mathcal{C}1}$.
Subsequently, during the \textbf{Atomic Reasoning Execution} stage, 
\model answers the original question by recursively executing the reasoning tree in bottom-up order.
For the execution of leaf atomic knowledge operators, \model dynamically selects, retrieves, and manipulates knowledge from multiple heterogeneous sources, effectively addressing challenge $\mathbf{\mathcal{C}2}$.
For other non-leaf nodes, the output is either produced by (1) Child Answer Reasoning, which formulates an answer based on its child nodes without additional retrieval, or (2) Direct RAG Reasoning, which is triggered only when Child Answer Reasoning fails and 
conducts retrieval
for the current node. 
This hierarchical execution process enhances the framework’s robustness and cost efficiency, ensuring that retrieval is only triggered when needed.

Finally, to address the lack of high quality datasets built on heterogeneous knowledge sources ($\mathbf{\mathcal{C}3}$), we construct a novel benchmark \textbf{\dataset} across three 
sources: Text, Web, and KG.
\dataset is constructed through a combination of LLM-generated content and careful manual verification.
We adopt a bottom-up construction approach to first create sub-questions within different knowledge sources, then merging them into one compositional question through a common bridging entity. 
For instance, in web pages, we collect recent news articles and leverage LLMs to ask questions about entities mentioned in the articles. 
This way, the data we construct cover a wide range of new and general knowledge, with minimal overlap between different knowledge sources. 
\dataset includes various types of questions such as multi-hop, comparison, true or false, and long answer. 
Previous methods only achieve as high as $33.85\%$ F1 scores on \dataset, showcasing its difficulty.

We 
evaluate \model on
three single-source and two multi-source knowledge-intensive QA datasets (including \dataset).
Experiments show that AtomR yields significant improvements over the previous SOTA ProbTree, achieving $5.4\%$, $9.4\%$, and $1.2\%$ F1 score improvements on single-source datasets, and $9.5\%$, $6.6\%$ F1 score improvements on multi-source datasets.
Furthermore, these improvements are achieved with less LLM and retriever calls.
Our contributions are three-fold:
(1) Introduce \model, an atomic operator-empowered LLM reasoning framework;
(2) Present benchmark \dataset built across three heterogeneous knowledge sources;
(3) Evaluate AtomR with extensive experiments and analysis, yielding new SOTA results on 3 single-source and 2 multi-source datasets. 

 \section{Related Work}

\paragraph{Knowledge-Intensive Reasoning.}

Knowledge-intensive reasoning requires both massive world knowledge and 
compositional reasoning skills.
It reflects a model's ability of using extensive knowledge to solve real-world problems: more specifically, to deduce accurate conclusions by effectively aggregating scattered pieces of knowledge.
Knowledge-intensive reasoning can be evaluated through a wide range of tasks, such as multi-hop question answering (QA)~\citep{hotpotqa, 2wiki, musique},
fact verification~\citep{DBLP:conf/naacl/ThorneVCM18, DBLP:conf/emnlp/WaddenLLWZCH20}, and knowledgeable open dialogue~\citep{DBLP:conf/iclr/DinanRSFAW19, DBLP:conf/kdd/YuZXLGZ0L022}. 
In this work, we focus on the multi-hop QA task for its effective modeling of complex reasoning structures.
Furthermore, the fact that world knowledge is dispersed across diverse structures and sources calls for knowledge-intensive reasoning systems to incorporate multiple heterogeneous knowledge sources.
Some recent works~\citep{DBLP:conf/iclr/LiZCDJPB24, DBLP:journals/tacl/0001BGV24} explore the integration of heterogeneous knowledge retrieval.
However, these works retrieve knowledge from the same static set of sources throughout each entire question, instead of dynamically retrieving from different suitable source(s) at each reasoning step, which is common in real-world scenarios.
We also observe that existing multi-source benchmarks~\citep{DBLP:conf/www/ChristmannRW24, crag} fail to assess such step-level cross-knowledge source retrieval.
Hence, we design \model to support flexible knowledge source selection at each atomic operator, and construct \dataset to evaluate cross-knowledge source reasoning.

\paragraph{Retrieval-Augmented Language Models.}

Retrieval-augmented LLMs have shown promising potential in knowledge-intensive reasoning.
Early works adopt a one-time retrieval strategy~\citep{atlas, DBLP:conf/icml/BorgeaudMHCRM0L22}, where the retriever is only called once to solve each question.
This 
is often insufficient for solving multi-hop questions, which requires retrieval at each reasoning step.
Hence, more recent approaches turn to multi-time retrieval strategies. 
IRCoT~\citep{DBLP:conf/acl/TrivediBKS23} and ITER-RETGEN~\citep{DBLP:conf/emnlp/ShaoGSHDC23} iteratively calls retrieval at each CoT step based on the previous steps' results. 
Self-Ask~\citep{DBLP:conf/emnlp/PressZMSSL23} iteratively 
generates
a chain of sub-questions and performs retrieval at each sub-question. 
ProbTree~\citep{DBLP:conf/emnlp/CaoZSL0THL23} proposes to first formulate a global question decomposition plan in tree structure, then recursively perform retrieval at each sub-question.
However, existing multi-time retrieval frameworks leverage LLMs to generate sub-questions in a free-form manner, imposing no supervision over each sub-question's atomicity and functionality.
Compared to previous methods, \model uses the combination of three predefined atomic knowledge operators to model complex reasoning procedures, achieving question decomposition that is highly fine-grained and orthogonal.

\section{Methodology}

\begin{figure*}[ht]
  \noindent
  \centering
  \scalebox{1.0}{
  \includegraphics[width=\textwidth]{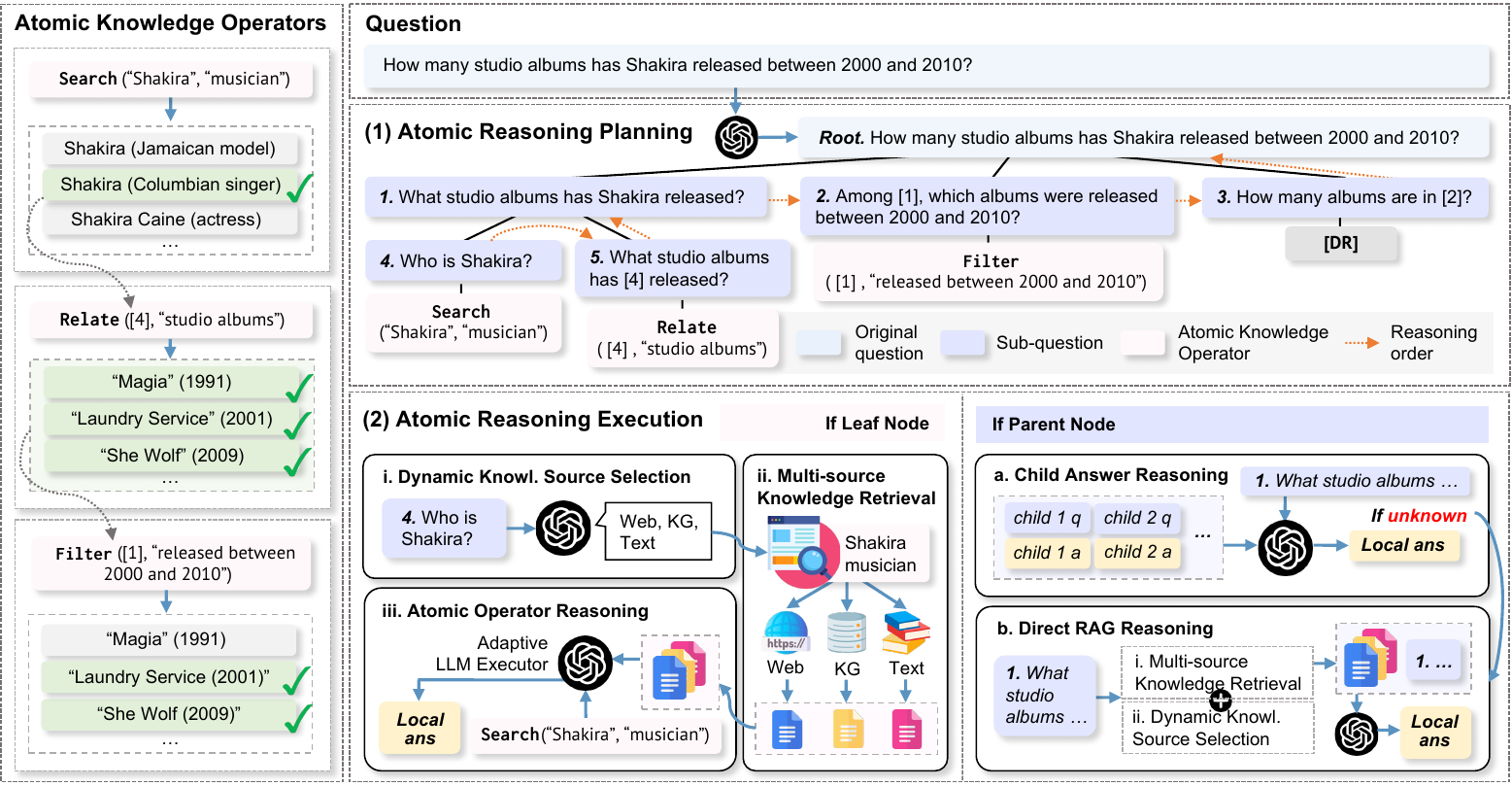}
  }
  \caption{The Overall Framework of AtomR.}
  \label{fig:system_architecture}
\end{figure*}

In this section, we introduce the design of \model.
The overall architecture of \model is illustrated in Figure~\ref{fig:system_architecture}. 
At the core of our framework are three fundamental \textbf{Atomic Knowledge Operators} (\Cref{sec:atomic_operators}), which we design to retrieve and manipulate knowledge at the atomic level.
These operators then steer the process of \model through two main stages: \textbf{(1) Atomic Reasoning Planning} (\Cref{sec:atomic_planning}), where the system 
decomposes the input question into a 
fine-grained Atomic Reasoning Tree (ART), and \textbf{(2) Atomic Reasoning Execution} (\Cref{sec:atomic_reasoning}), where the system performs bottom-up reasoning across multiple heterogeneous knowledge sources at each atomic node. 
We first introduce the design of our 
atomic knowledge operators in Section~\ref{sec:atomic_operators}, then detail the two main stages of our framework in Sections~\ref{sec:atomic_planning} and~\ref{sec:atomic_reasoning}.
The LLM prompts of \model are included in Appendix~\ref{sec:prompts}.
The benchmark construction of \dataset will be introduced later in Section~\ref{sec:blendqa}.

\subsection{Atomic Knowledge Operators}
\label{sec:atomic_operators}

\begin{table}[]
\caption{The correspondence of each \model atomic knowledge operator to essential SPARQL and Cypher clauses and KoPL functions. ``[REL]'' represents inter-nodal relationship in Cypher, while ``WHERE(n)'' and ``WHERE(e,r)'' represents using SPARQL's WHERE clause with the entity name constraint and the entity-relation constraint, respectively. 
}
\label{tab:atomic_operator_mapping}
\scalebox{0.92}{
\begin{tabular}{>{\raggedright\arraybackslash}m{1.7cm}>{\centering\arraybackslash}m{1.1cm}>{\centering\arraybackslash}m{1.1cm}>{\centering\arraybackslash}m{2cm}>
{\centering\arraybackslash}m{1.05cm}}
    \toprule
      Function & SPARQL  & Cypher  & KoPL & \textbf{\model} \\
    \midrule 
    Entity Disambiguation 
        & WHERE (n)
        & MATCH  
        & Find 
        &  \textbf{\search} \\
    \midrule 
    One-hop Inference
        & WHERE (e,r)
        & [REL]
        & Relate, QueryAttr, QueryRelation  & \textbf{\relate} \\ 
    \midrule 
    Entity Filtering 
        & FILTER
        & WHERE 
        & FilterConcept, FilterStr, FilterNum, FilterYear, FilterDate  
        & \textbf{\filter} \\ 
    \bottomrule
\end{tabular}
}
\end{table}

Compositional reasoning, the ability to construct complex reasoning procedures by combining simple components, is a fundamental property of human thought~\citep{fodor1988connectionism, fodor2002compositionality}.
Knowledge graphs (KGs) effectively model the compositionality of knowledge, breaking down complex world knowledge into combinations of atomic elements: entities, relations, and attributes.
Furthermore, KG query languages use predefined operations to manipulate the atomic elements in KGs.
In this light, we ask the question: 
\textit{Can we use a set of predefined operators to atomically manipulate knowledge and enhance LLM compositional reasoning?}

We investigate the essential operations of existing KG query languages, and observe that all operations can be distilled into combinations of three basic \textit{atomic knowledge operators}: \textbf{\search}, \textbf{\relate}, and \textbf{\filter}.
Table~\ref{tab:atomic_operator_mapping} illustrates how operations of three widely-used KG query languages, SPARQL, Cypher, and KoPL~\citep{DBLP:conf/semweb/PerezAG06, DBLP:conf/sigmod/FrancisGGLLMPRS18, DBLP:conf/acl/CaoSPNX0LHZ22}, can be induced into \model's three atomic knowledge operators.
Each operator corresponds to a unique, orthogonal function.
We implement each operator on multiple heterogeneous knowledge sources with an adaptive LLM executor, as shown in Figure~\ref{fig:system_architecture}(2)(iii).

\paragraph{\textbf{\texttt{Search}}}
The \search operator is designed for \textit{entity disambiguation}: to accurately retrieve the desired entities from a massive entity pool.
\model's \search operator is defined as follows:
\begin{equation*}
    list[entity] = \mathtt{Search}(entity\_name, \{optional\} descriptor) \nonumber
\end{equation*}
It contains two inputs: the entity name and an optional entity descriptor to facilitate disambiguation.
For example, \search\textit{(``Michael Jordan'', ``scientist'')} returns the entity \textit{[``Michael I. Jordan'']}, successfully disambiguating the machine learning scientist ``Michael Jordan'' from other ``Michael Jordan'' enities.

During execution, \search first initiates multi-source knowledge retrieval using concatenated parameters \textit{``\{entity\_name\} \{descriptor\}''} as the query, which will be detailed in Section~\ref{sec:multi_source_retrieval}.
Then, the retrieved knowledge and the operator's corresponding sub-question are inputted to an \textit{adaptive LLM executor} to conduct entity disambiguation through in-context learning (Figure~\ref{fig:prompt_search}). Finally, the adaptive LLM executor outputs the disambiguated entity list.

\paragraph{\textbf{\texttt{Relate}}}

The \relate operator is designed for \textit{one-hop inference}.
There are three possibilities for \relate: (1) retrieve tail entities given a head entity and relation, (2) retrieve an attribute value given an entity and attribute, and (3) retrieve a relation between a head and tail entity.
The three possibilities are depicted in Equations~\ref{eq:e_r},~\ref{eq:e_a},~\ref{eq:e_e}, where $entity_h$ and $entity_t$ denote head and tail entity:

\begin{align}
    list[entity_{t}] &= \relate(entity_{h}, relation) \label{eq:e_r} \\
    value &= \relate(entity_{h}, attribute) \label{eq:e_a}\\
    relation &= \relate(entity_{h}, entity_{t}) \label{eq:e_e}
\end{align}

An example of case (1) would be \textit{Relate(``Barack Obama'', ``child'')}, which returns the tail entity list \textit{[``Malia Obama'', ``Sasha Obama'']}. 
Similarly, for case (2), \textit{Relate(``Barack Obama'', ``date of birth'')} returns the attribute \textit{[``August 4th, 1961'']}.
An example for case (3) would be to reversely retrieve the relation of Malia Obama to Barack Obama, where \textit{Relate(``Malia Obama'', ``Barack Obama'')} returns \textit{[``child'']}.

During execution, similar as \search, \relate first initiates multi-source knowledge retrieval using the concatenated parameters as the query,
then inputs the retrieved knowledge and corresponding sub-question into the adaptive LLM executor with an in-context learning prompt.
Finally, the LLM executor outputs the answer: either an entity list for case (1), an attribute value for case (2), or a relation for case (3).

\paragraph{\textbf{\texttt{Filter}}}

The \filter operator is designed for \textit{entity filtering} based on an attributal condition.
It is defined as follows: 
\begin{align}
    list[entity] &= \filter(list[entity], condition)  \nonumber
\end{align}
For example, \textit{Filter([``Lionel Messi'', ``Steven Jobs'', ``Bill Gates''], ``born in 1955'')}, returns \textit{[``Bill Gates'', ``Steve Jobs'']}.

The execution of \filter is slightly more complicated.
First, \filter initiates multi-source knowledge retrieval for every entity in its input $list[entity]$, where each retrieval query is formulated as \textit{``\{entity\_name\} \{condition\}''}.
Subsequently, for each entity $e_i$, \filter concatenates all retrieved passages and calculates the overlapping coefficient $O_i$ between the query $q_i$ \textit{``\{entity\_name\} \{condition\}''} and the concatenated passage $p_i$:

\begin{align}
    O_{i}(q_i, p_i) = \frac{|q_i \cap p_i|}{\min(|q_i|, |p_i|)}  \nonumber
\end{align}

The overlapping coefficient of each entity is then compared to a hyperparameter threshold $t$.
Entities with $O_{i}<t$ will be directly discarded to remove excessive noise from the retrieved knowledge.
Then, the remaining entities and passages are inputted to the adaptive LLM executor to perform entity filtering via in-context learning.
Finally, the LLM executor outputs the filtered entity list.

\subsection{Atomic Reasoning Planning}
\label{sec:atomic_planning}

To 
model the complex structures of knowledge-intensive reasoning, \model employs reasoning planning in a tree structure \citep{DBLP:conf/nips/YaoYZS00N23, DBLP:conf/emnlp/CaoZSL0THL23}.
Given a complex question, \model leverages an LLM planner 
to decompose the question into an \textit{Atomic Reasoning Tree} (ART), where the root node is the original complex question, while each non-root node is a decomposed sub-question of its parent. 
The decomposition continues until each leaf question is a fine-grained \textit{atomic question}, which could be directly answered by either (a) calling one of the three 
atomic knowledge operators—\search, \relate, or \filter, or (b) analyzing answers of previous sibling sub-questions.
Then, for each 
atomic 
question of case (a), the LLM planner appends a corresponding atomic operator as the final set of leaf nodes for the ART.
We generate ARTs using in-context learning (Figure~\ref{fig:prompt_ART}).

Following Cao et al. \citep{DBLP:conf/emnlp/CaoZSL0THL23}, we index ART nodes in breadth-first search (BFS) order. 
To reference intermediate answers, we use reference placeholders denoted as $[i]$, which will be substituted with actual answers of sub-question $[i]$ during the reasoning execution stage.
Figure~\ref{fig:system_architecture}(1) illustrates an example ART, where the original question ``How many studio albums has Shakira released between 2000 and 2010?'' is decomposed into a reasoning tree of 3 leaf atomic knowledge operators, linked to sub-questions 4, 5, and 2, respectively. 
The grey \textit{[DR]} (Direct Reasoning) mark attached to sub-question 3 ``How many albums are in \textit{[2]}?'' indicates that it does not trigger a leaf operator call but instead formulates an answer by direct LLM reasoning based on the output of its previous sibling sub-questions, in this case sub-question 2.

We argue that an ART is \textit{atomic} because a question is decomposed until each sub-question reaches an atomic granularity that could be handled by a predefined atomic knowledge operator.
Although previous efforts have sought to achieve fine-grained question decomposition, they often fall short due to the lack of effective atomic constraints.
By explicitly inducing and defining a set of atomic knowledge operators, \model successfully manoeuvres the LLM's reasoning planning process at a highly fine-grained level.

\subsection{Atomic Reasoning Execution}
\label{sec:atomic_reasoning}

In this stage, given the decomposed ART, we conduct bottom-up reasoning over each node via post-order-traversal.
Figure~\ref{fig:system_architecture}(2) visualizes the main modules and reasoning procedure of the Atomic Reasoning Execution stage.
Due to space limitations, the pseudo-code of the detailed algorithm is shown in Appendix~\ref{sec:alg_reasoning_execution}.

\subsubsection{\textbf{Leaf Node Reasoning}}

In an ART, each leaf node is an atomic knowledge operator.
The execution of an atomic knowledge operator consists of three steps: (1) Dynamic knowledge Source Selection, (2) Multi-source knowledge retrieval, and (3) Atomic Operator Reasoning. 

\subparagraph{\textbf{Dynamic Knowledge Source Selection.}}

While previous works retrieve knowledge from a static knowledge source, \model employs dynamic knowledge source selection at each sub-question.
This enables \model to flexibly identify the most suitable knowledge sources to answer each sub-question.
Specifically, we leverage an LLM as a dynamic knowledge source selector through in-context learning.
The LLM outputs the selected knowledge sources, which are used in the subsequent Multi-source Knowledge Retrieval step.

\subparagraph{\textbf{Multi-source Knowledge Retrieval.}}
\label{sec:multi_source_retrieval}

After selecting the appropriate knowledge sources for each sub-question, \model initiates knowledge retrieval from each source.
The retrieval query and method varies by source.
In terms of retrieval query, web and text sources use the queries formulated in Section~\ref{sec:atomic_operators}, while KG sources use an external semantic parser to parse the sub-question into a structured query program.
In terms of retrieval method, web sources are queried via a search engine API, text sources via a dense retriever, and KGs via a structured query language. 
For the text and web knowledge sources, \model retrieves the top $k$ passages and articles, where $k$ is a hyperparameter.
For the KG knowledge source, the full 
answer list is returned as a serialized string.

\subparagraph{\textbf{Atomic Operator Reasoning.}}

Finally, with the retrieved knowledge, \model executes the atomic knowledge operator.
The detailed functionality and procedures of each atomic knowledge operator has been explained in Section~\ref{sec:atomic_operators}.
To ensure flexibility and robustness, we utilize an \textit{Adaptive LLM Executor}, instead of static symbolic code, for atomic operator execution.
Specifically, we employ in-context learning to prompt an LLM to perform the desired knowledge operation (i.e., Figure~\ref{fig:prompt_search}).
Finally, the output of the adaptive LLM executor serves as the local answer for the atomic operator as well as its associated sub-question.

\subsubsection{\textbf{Parent Node Reasoning}}

In an ART, all nodes that don't initiate an atomic knowledge operator call are considered parent nodes.
A parent node may go through (1) Child Answer Reasoning, (2) Sibling Anwer Reasoning, and (3) Direct RAG Reasoning.

\subparagraph{\textbf{Child Answer Reasoning.}}

Child answer reasoning is the process of deducing an answer for a parent node by synthesizing answers of its child nodes, formally:
\begin{equation*}
a_i = ChildAnswerReasoning(q_i, [q_{c_1}, a_{c_1}, q_{c_2}, a_{c_2}, ...]) \nonumber
\end{equation*}
where $q_i$ is the sub-question for the current node and $[q_{c_1}, a_{c_1}, ...]$ is the list of question-answer pairs for $q_i$'s child nodes.
For example, in Figure~\ref{fig:system_architecture}(1), sub-question 1 can be answered by synthesizing the answers of its child nodes 4 and 5. 
Child answer reasoning is achieved through in-context learning, where the LLM is provided with the child question-answer pairs along with an instruction.

\subparagraph{\textbf{Sibling Answer Reasoning.}}

Similar to child answer reasoning, sibling answer reasoning targets \textit{[DR]} nodes that are answered by analyzing answers of its previous sibling nodes, formally: 
\begin{align}
    a_i = SiblingAnswerReasoning(q_i, [q_{s_1}, a_{s_1}, q_{s_2}, a_{s_2}, ...]) \nonumber
\end{align}
where $[q_{s_1}, a_{s_1}, ...]$ is the list of question-answer pairs for $q_i$'s sibling nodes that are referenced in the current question $q_i$.
For example, in Figure~\ref{fig:system_architecture}(1), sub-question 3 can be answered by analyzing answers of its referenced sibling node 2.
Sibling answer reasoning is also achieved through LLM in-context learning.

\subparagraph{\textbf{Direct RAG Reasoning.}}

Each parent node is primarily resolved through either child answer reasoning or sibling answer reasoning.
However, if a parent node fails to obtain an answer, \textit{direct RAG reasoning} is employed.
Parallel to leaf node reasoning, direct rag reasoning first performs dynamic knowledge source selection for the current sub-question $q_i$, then initiates multi-source knowledge retrieval with $q_i$ as the query.
Finally, given $q_i$ and the retrieved knowledge, an LLM is employed to conduct standard RAG through in-context learning.
This design ensures that while \model primarily focuses on knowledge retrieval at the leaf nodes, it can also flexibly access external knowledge at inner nodes when necessary.

\section{Experiments}

\begin{table*}[t]
\caption{Results for Single-Source Reasoning. Token-level F1 is reported, with the best results in bold and the second best results \underline{underlined}. \textcolor{darkgreen}{$(\uparrow)$} indicates the overall gain of \model compared to the second-best baseline for each dataset.}
\label{tab:main_results_single}
\scalebox{0.95}{
\begin{tabular}{lcccccccccccc}
    \toprule
    & \multicolumn{3}{c}{HotpotQA}  & \multicolumn{5}{c}{2Wikimultihop}   & \multicolumn{4}{c}{Musique} \\
    \cmidrule(lr){2-4}  \cmidrule(lr){5-9}  \cmidrule(lr){10-13}
    & Overall & Bridge & Comp. & Overall & Bridge & Infer. & Comp. & B.C. & Overall & 2hop & 3hop & 4hop \\
    \midrule
    \multicolumn{3}{l}{\textit{Without Retrieval (Closebook)}} \\
    \midrule
    Standard Prompting &  50.02       &  45.22      &  72.50      &  41.09       &  14.44      &  32.24         &  71.63          & 63.13     &  19.31       &  21.84    &  16.03    &  17.81    \\
    CoT              &  58.38  &  55.48   & 71.91  &  58.32  & 38.24  & 54.83   &  77.27   &  77.17    & 29.03   &  36.26    & 24.07     &  17.38    \\
    \midrule
    \multicolumn{3}{l}{\textit{With Retrieval (Wikipedia)}}  \\
    \midrule
    Standard RAG &  60.31  & 61.88   & 52.97   &  47.94       &  34.96      &  54.32         &  56.96      &  57.27    &   22.07      &  28.51    &  15.81    & 14.77     \\
    IRCoT      & 60.20  & 58.00  & \underline{69.40}  & 63.80  & 46.20  & 45.50  & \underline{91.60}  & 79.00  & 34.20  & \underline{44.20}  & 26.30  & 20.10    \\
    SearChain        & 59.04  & \textbf{69.73}  & 51.12  & 63.10  & 48.85  & 50.38  & 81.41  & 84.21  & 31.68  & 38.89  & 28.38  & 17.28  \\
    ProbTree         & \underline{65.91}  & 65.50  & 67.81  & \underline{69.32}  & \underline{52.45}  & \underline{64.08}  & 88.17  & \underline{90.00}  & \underline{34.92}  & 42.52  & \textbf{30.38}  & \textbf{21.53}   \\
    \midrule
    \textbf{\model (ours)}     & \textbf{71.27{\footnotesize \textcolor{darkgreen}{$(\uparrow5.4)$}}}  & \underline{68.96}  & \textbf{82.07}       & \textbf{78.72{\footnotesize \textcolor{darkgreen}{$(\uparrow9.4)$}}}  & \textbf{59.07}  & \textbf{80.04}  & \textbf{97.48}    & \textbf{93.33} & \textbf{36.11{\footnotesize \textcolor{darkgreen}{$(\uparrow1.2)$}}}  & \textbf{45.63}  & \underline{29.94} & \underline{20.15} \\
    \bottomrule
\end{tabular}
}
\end{table*}
\begin{table*}[t]
\caption{Results for Multi-Source Reasoning. Token-level F1 is reported. Web, Text, and KG denote the integration of knowledge from Google, the Wikipedia corpus, and the Wikidata knowledge base, respectively.}
\label{tab:main_results_multi}
\scalebox{0.95}{
\begin{tabular}{lC{0.65cm}C{0.65cm}C{0.65cm}ccccccccc}
    \toprule
    & \multirow{2}{*}{Web} & \multirow{2}{*}{Text} & \multirow{2}{*}{KG} & \multicolumn{4}{c}{BlendQA} & \multicolumn{5}{c}{CRAG}   \\
    \cmidrule(lr){5-8}  \cmidrule(lr){9-13}  
    & & & & Overall & KG-Web & KG-Text & Text-Web & Overall & Simple & S.C. & Comp. & M.H. \\
    \midrule
    \multicolumn{3}{l}{\textit{Without Retrieval (Closebook)}} \\
    \midrule
    Standard Prompting & - & - & - & 23.26  & 16.49  & 24.71  & 27.64  & 59.18  & 56.46  & 53.87  & 70.52  & 55.86   \\
    CoT  & - & - & - & 30.61  & 18.30  & 34.37  & 37.24 & 63.59  & 63.22  & 60.09  & 69.53  & 61.40   \\
    \midrule
    \multicolumn{3}{l}{\textit{With Retrieval (Multi-Source)}} \\
    \midrule
    Standard RAG & \checkmark  & \checkmark  & \checkmark & 33.78  & \underline{26.62}  & 32.76  & \underline{41.21}  & 59.27  & \underline{66.23}  & 56.70  & 56.67  & 56.98  \\
    Self-Ask & \checkmark & - & - & 26.25  & 20.69  & 27.59  & 29.68  & 48.40  & 47.52  & 43.22  & 60.27  & 42.30 \\
    ProbTree  & \checkmark  & \checkmark  & - & \underline{33.85}  & 24.48  & \underline{38.81}  & 36.70  & 58.78  & 58.95  & 49.00  & 67.90  & 59.28    \\
    Chain-of-Knowledge & \checkmark  & \checkmark  & \checkmark  & 33.08  & 22.66  & 37.55  & 37.38  & \underline{64.75}  & 57.82  & \underline{62.15}  & \underline{74.04}  & \textbf{65.40} \\
    \midrule
    \textbf{\model (ours)} & \checkmark  & \checkmark  & \checkmark  & \textbf{43.32{\footnotesize \textcolor{darkgreen}{$(\uparrow9.5)$}}}  & \textbf{45.63}  & \textbf{39.69}  & \textbf{45.22}  & \textbf{71.39{\footnotesize \textcolor{darkgreen}{$(\uparrow6.6)$}}}  & \textbf{70.10}  & \textbf{68.51}  & \textbf{81.71}  & \underline{64.95}  \\
    \bottomrule
\end{tabular}
}
\end{table*}

\subsection{Experimental Setup}

\subsubsection{\textbf{Datasets and Evaluation Metrics}}

We evaluate \model in both single-source and multi-source settings.
In the single-source setting, we evaluate \model on three Wikipedia-based multi-hop QA benchmarks: HotpotQA \citep{hotpotqa}, 2WikiMultiHop \citep{2wiki}, and Musique \citep{musique}. 
We adopt the test and development sets released by IRCoT \citep{DBLP:conf/acl/TrivediBKS23}, which include a 500-entry test set and a 100-entry development set.
In the multi-source setting, we evaluate \model on two recent datasets: CRAG \citep{crag} by Meta and \dataset by us.
Both CRAG and \dataset are constructed on three heterogeneous knowledge sources—the Wikidata knowledge base, the Wikipedia text corpus, and the Google Web search engine.
We will provide a more thorough introduction of \dataset in the next section.
For \dataset, we evaluate \model on the full 445-entry dataset;
for CRAG, we only consider static questions and sample a 500-entry test set with a 100-entry development set.
Following IRCoT and ProbTree, we adopt token-level F1 as the evaluation metric.

\subsubsection{\textbf{\dataset}}

\label{sec:blendqa}

We propose \dataset, a dataset to evaluate models' cross-knowledge source reasoning capabilities.
Specifically, we construct \dataset on three heterogeneous knowledge sources: a full Wikipedia dump as the text source, Google as the web source, and Wikidata as the knowledge graph.
We use \textit{gpt-4o-2024-08-06} as the LLM to aid dataset construction.
The general construction process is to generate two sub-questions $sub$-$q_1$ and $sub$-$q_2$ from two different sources that share a common bridging entity, and then merge them together to form a cohesive question. 
There are three types of questions: KG-Text, KG-Web, Text-Web. 

\noindent 
\textbf{KG-Text}. For KG-Text questions, $sub$-$q_1$ is sampled from NaturalQuestions (NQ)~\citep{DBLP:journals/tacl/KwiatkowskiPRCP19}, which is built on Wikipedia. 
We first identify $sub$-$q_1$'s topic entity as the bridging entity, then sample the topic entity's triples from the KG to generate $sub$-$q_2$.

\noindent 
\textbf{KG-Web}. There are two sub-types of KG-Web questions depending on the anchor of the constructed data. 
(1) \textit{KG2Web} is anchored in the KG. Specifically, we randomly sample entities from the KG as bridging entities, and sample their triples to generate $sub$-$q_1$.
Then, we search the bridging entity for relevant news to generate $sub$-$q_2$. 
(2) \textit{Web2KG} is anchored in the web. 
Specifically, we collect a large number of news articles from various domains (politics, business, etc.), and ask the LLM to identify a bridging entity in each article and ask a question about it ($sub$-$q_1$). 
Then, we generate $sub$-$q_2$ by sampling triples from the KG using the bridging entity.

\noindent 
\textbf{Text-Web}. There are also two sub-types of Text-Web questions. 
(1) \textit{Web2Text}. We sample $sub$-$q_1$ from NQ and identify the topic entity as the bridging entity.
Then, we search the entity on the web (excluding Wikipedia articles) to collect its descriptions. 
With these descriptions, we use the LLM to generate a unique tag of the entity to serve as $sub$-$q_2$, e.g. ``Neil Armstrong'' - ``first man to walk on the moon''.
(2) \textit{Text2Web}. Again, we sample $sub$-$q_1$ from NQ, and use the answer of it as the bridging entity (if valid). Then we search the entity's news articles to generate $sub$-$q_2$.

Overall, \dataset consists of 445 questions, with 132 KG-Web questions, 163 Text-KG questions, and 150 Text-Web questions.

\subsubsection{\textbf{Baselines}} 

\subparagraph{\textbf{Single-source Baselines.}}
We evaluate both Closebook and Openbook baselines.
For the Closebook setting, we evaluate Standard Prompting and CoT using two-shot prompts.
For the Openbook setting, we first evaluate Standard RAG with one-time retrieval, then evaluate three state-of-the-art reasoning frameworks: two chain-structured frameworks IRCoT \citep{DBLP:conf/acl/TrivediBKS23} and SearChain \citep{DBLP:conf/www/XuPSCC24}, and one tree-structured framework ProbTree \citep{DBLP:conf/emnlp/CaoZSL0THL23}.

\subparagraph{\textbf{Multi-source Baselines.}}
We evaluate both Closebook and Openbook approaches using Standard Prompting, CoT, and Standard RAG with two-shot prompts. 
We also assess three established frameworks: Self-Ask \citep{DBLP:conf/emnlp/PressZMSSL23}, which integrates only a Google search engine; ProbTree \citep{DBLP:conf/emnlp/CaoZSL0THL23}, which combines a Google engine with a Wikipedia corpus; and Chain-of-Knowledge (CoK) \citep{DBLP:conf/iclr/LiZCDJPB24}, which supports all three knowledge sources—Google, Wikipedia, and Wikidata.

\subsubsection{\textbf{Implementation Details}}

For all experiments, we use \textit{gpt-4o-2024-08-06} as our base LLM.
For web knowledge retrieval, we use SERPAPI to access Google.
For text knowledge retrieval, we follow previous works~\citep{mcr, DBLP:conf/www/XuPSCC24} to use ColBERTv2 as our text retriever.
We use HotpotQA's official Wikipedia abstract dump from October 2017 to evaluate HotpotQA, and the Dec 2021 full Wikipedia dump from Atlas~\citep{atlas} to evaluate the other four datasets.
For KB knowledge retrieval, we use KoPL as the query language and KQA Pro~\citep{DBLP:conf/acl/CaoSPNX0LHZ22}'s Wikidata dump as the knowledge base.
We only retrieve text knowledge for single-source datasets, and retrieve knowledge from as many sources as supported for each baseline on multi-source datasets.
We retrieve knowledge from all three sources for multi-source Standard RAG.
For \model, we set $k=3$ to retrieve the top 3 web and text results, and set $t=0.5$ for the \filter function.
The results for IRCoT are from Cao et al. \citep{DBLP:conf/emnlp/CaoZSL0THL23}, while all other baselines are reproduced by ourselves using the above settings.

\subsection{Main Results}

\subsubsection{\textbf{Single-source Results}} 

As shown in Table~\ref{tab:main_results_single}, \model outperforms all baselines across all three datasets.
Compared to the previous SOTA ProbTree, \model achieves F1 improvements by $5.4\%$, $9.4\%$, and $1.2\%$ on HotpotQA, 2WikiMultiHop, and Musique, respectively.
\model also demonstrates outstanding performance on all question types,
achieving F1 scores as high as $97.48\%$ and $93.33\%$ on the Comparison and Bridge-Comparison question types.
Such improvements demonstrate the effectiveness of \model's atomic knowledge operators to model complex reasoning processes.

While \model achieves the best overall performance on Musique, it yields slightly lower results on 3-hop and 4-hop questions.
Upon closer investigation,
we find that gold sub-questions annotated in Musique are relatively coarse-grained and can usually be decomposed into 2 to 3 sub-questions by \model. 
As a result, \model may inadvertently over-decompose 3-hop and 4-hop questions.

\subsubsection{\textbf{Multi-source Results}}

\model achieves the best results on both \dataset and CRAG, with overall F1 improvements of $9.5\%$ and $6.6\%$, respectively.
On \dataset, \model outperforms all baselines consistently over each knowledge-source setting, demonstrating the effectiveness of our dynamic knowledge selection process. 
We also observe that compared to \model, the two previous SOTA frameworks ProbTree and Chain-of-Knowledge are more prone to LLM hallucination because they rely heavily on the LLM's memorized factual knowledge during the reasoning process.
In contrast, \model prioritizes reasoning with accurately retrieved fine-grained knowledge, leading to more trustworthy results.
Furthermore, comparing to ProbTree, \model achieves more superior results with fewer LLM and retriever calls, which is detailed in Section~\ref{sec:cost_analysis}.

\subsection{Analysis}

\subsubsection{\textbf{Comparative Case Study: How does \model outperform previous methods?}}

\begin{figure*}[!ht]
  \noindent
  \centering
  \scalebox{0.95}{
  \includegraphics[width=\linewidth]{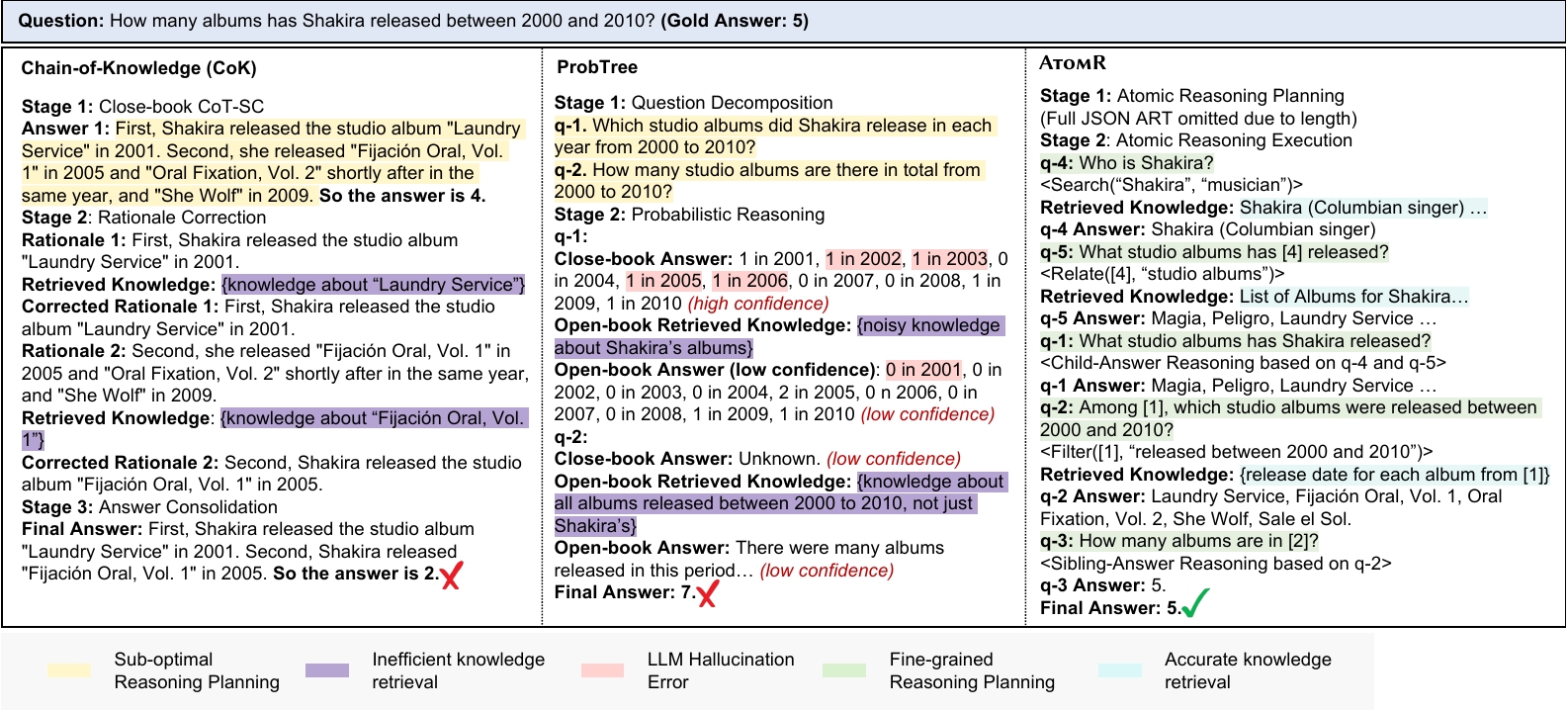}
  }
  \caption{Comparative case study of \model with previous SOTA Chain-of-Knowledge and ProbTree.}
  \Description{}
  \label{fig:case_study}
\end{figure*}

In Figure~\ref{fig:case_study}, we present a specific example to showcase three advantages of \model compared to two previous SOTA frameworks - ProbTree and CoK.

\subparagraph{\textbf{(1) Effective Reasoning Planning.}}

In this test case, CoK decomposes the question into two sequential rationales, each recalling only a few of Shakira's studio albums, leading to an incomplete answer. 
ProbTree decomposes the input into two overlapping complex sub-questions, resulting in computational redundancy and inaccurate results.
In contrast, \model achieves complete, fine-grained, and orthogonal reasoning planning.

\subparagraph{\textbf{(2) Accurate Knowledge Retrieval.}}
The sub-optimal reasoning planning of CoK and ProbTree directly leads to inefficiencies in knowledge retrieval. 
CoK only retrieves information of the albums mentioned in each rationale, capturing just a subset of the required answer set. 
ProbTree’s compositional sub-questions result in noisy and inaccurate retrievals.
In contrast, \model's atomic reasoning planning ensures fine-grained and precise knowledge retrieval.

\subparagraph{\textbf{(3) Preventing LLM hallucination.}}
CoK and ProbTree rely heavily on LLM parametric knowledge, both prioritizing LLM close-book answering.
This makes both frameworks susceptible to LLM hallucination. 
For instance, ProbTree’s close-book answer for q-1 contains 4 factual errors.
In contrast, \model uses LLMs as reasoning agents and prioritizes accurately retrieved knowledge.

\subsubsection{\textbf{Effect of Incorporating Multi-Source Knowledge}}

\begin{table}[t]
\caption{Ablating the integration of knowledge sources.}
\label{tab:knowledge_sources}
\scalebox{0.9}{
\begin{tabular}{lcc}
    \toprule
      Knowledge Sources & \dataset  & CRAG \\
    \midrule 
    Text, Web, KG  & \textbf{46.97}  & \textbf{70.23}  \\
    \midrule 
    w/o KG  &  46.52\textcolor{darkred}{$(\downarrow0.5)$} & 68.88\textcolor{darkred}{$(\downarrow1.4)$}  \\ 
    w/o Text & 43.34\textcolor{darkred}{$(\downarrow3.6)$} & 67.03\textcolor{darkred}{$(\downarrow3.2)$} \\ 
    w/o Web & 30.30\textcolor{darkred}{$(\downarrow16.7)$} & 58.27\textcolor{darkred}{$(\downarrow12.0)$} \\
    \bottomrule
\end{tabular}
}
\end{table}

We conduct an ablation study to assess the impact of multi-source knowledge integration by sampling 80 entries each from \dataset and CRAG. 
The results, detailed in Table~\ref{tab:knowledge_sources}, indicate that removing any knowledge source reduces \model's performance, with Web knowledge having the most significant impact and KG the least. 
Removing the Text source results in a noticeable decline of over $3\%$, highlighting that Wikipedia corpus retrieval is necessary and cannot be fully substituted by Web searches. 
We also report the test-time distribution of selected knowledge sources for \model in Figure~\ref{fig:distribution_knowledge_sources} of Appendix~\ref{sec:distribution_knowledge_sources}.

\subsubsection{\textbf{Comparative Cost Analysis}}
\label{sec:cost_analysis}

\begin{figure}[ht]
  \noindent
  \centering
  \includegraphics[width=\linewidth]{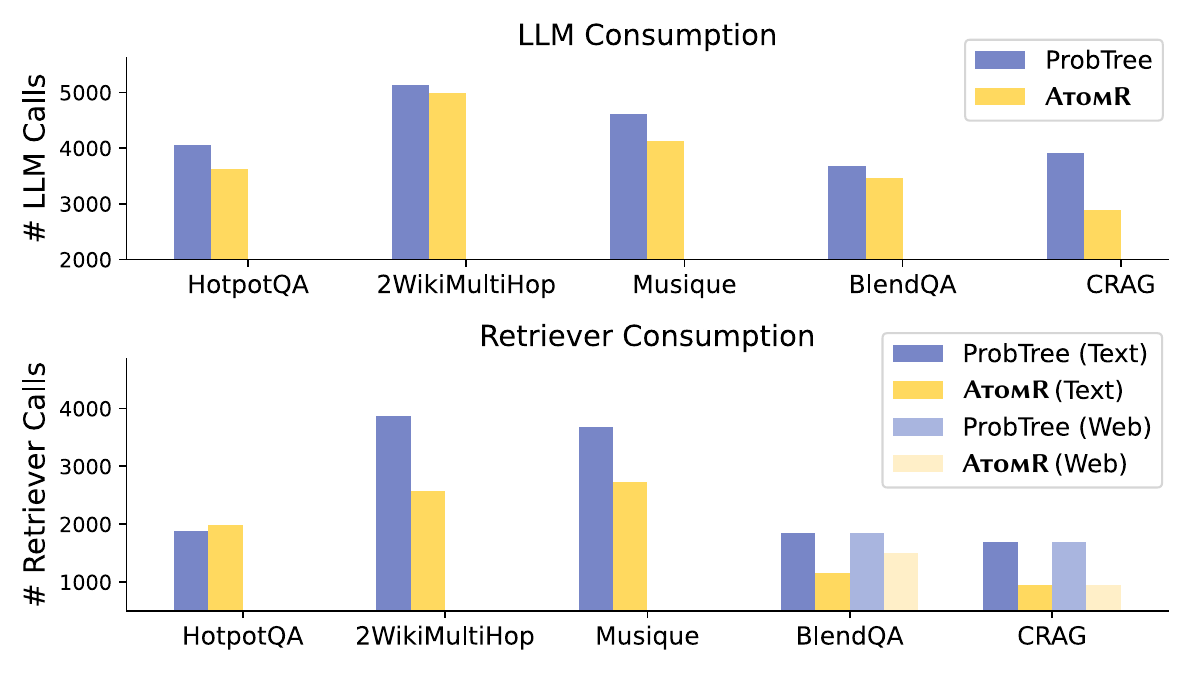}
  \caption{Comparing \model's consumption costs with the previous SOTA ProbTree.}
  \Description{}
  \label{fig:api_consumption}
\end{figure}

Both \model and the previous SOTA framework ProbTree employ tree reasoning structures. 
Yet, we design \model to be more accurate while also more \textit{cost-efficient}.
Figure~\ref{fig:api_consumption}
illustrates that \model requires fewer LLM and retriever calls compared to ProbTree. 
For LLM consumption, ProbTree necessitates at least two LLM calls per tree node: one for close-book answering and another for open-book answering.
In contrast, \model demands just one LLM call per node, whether for Atomic Operator Reasoning at a leaf node or for Child-or-Sibling Answer Reasoning at a parent node. 
Similarly, while ProbTree triggers at least one retriever call at each node, \model only initiates retriever calls when external knowledge is needed--either at an atomic leaf node, or at a parent node that fails to formulate an answer with Child Answer Reasoning.
This is made possible by the \textit{atomic} nature of our method: \model decomposes complex questions into fine-grained, orthogonal sub-questions, 
allowing retrieval to be limited to leaf nodes in most cases.

To further evaluate the effectiveness of \model, we include more analysis and ablations in the Appendix: the distribution of atomic knowledge operator calls (Appendix~\ref{sec:distribution_operators}), the distribution of selected knowledge sources (Appendix~\ref{sec:distribution_knowledge_sources}), backbone LLM ablations (Appendix~\ref{sec:ablating_llms}), and text retriever ablations (Appendix~\ref{sec:ablating_retrievers}).
\section{Conclusion}

In this paper, we propose \model, an Atomic operator-empowered LLM reasoning framework for heterogeneous knowledge sources. \model breaks down the reasoning process to the atomic level of knowledge, using the combination of three fundamental atomic knowledge operators to model complex reasoning procedures.
We also propose \dataset, a benchmark evaluating cross knowledge-source reasoning.
Experiments on three single-source and two multi-source 
datasets demonstrate that \model outperforms previous SOTA frameworks by large margins, showcasing its superiority.

\begin{acks}
This work is supported by National Natural Science Foundation of China (62476150), Beijing Natural Science Foundation (L243006), and Tsinghua University Initiative Scientific Research Program.
\end{acks}

\bibliographystyle{ACM-Reference-Format}
\bibliography{custom}

\appendix

\section{Algorithm for Atomic Reasoning Execution}
\label{sec:alg_reasoning_execution}

Algorithm~\ref{alg:reasoning_execution} depicts the pseudo-code of the Atomic Reasoning Execution stage detailed in Section~\ref{sec:atomic_reasoning}.

\begin{algorithm}[htp]
\algblock[TryCatchFinally]{try}{endtry}
\algcblock[TryCatchFinally]{TryCatchFinally}{finally}{endtry}
\algcblockdefx[TryCatchFinally]{TryCatchFinally}{catch}{endtry}
	[1]{\textbf{catch} #1}
	{\textbf{end try}}
\caption{Atomic Reasoning Execution}
\begin{algorithmic}[1] 
\State \textbf{Require:} Atomic Reasoning Tree $ART$
\For{$n_i$ in PostOrder($ART$)}
    \If{$n_i$ is leaf}
        \State \textbf{prepare} $n_i$'s sub-question $q_i$, child $c_i$
        \State $s_i \gets KnowledgeSourceSelection(q_i)$
        \State $k_i \gets KnowledgeRetrieval(c_i, s_i)$
        \try
            \State $a_i \gets AtomicOperatorReasoning(q_i, c_i, k_i)$
        \catch{Operator Execution Failure}
            \State $a_i \gets DirectRAGReasoning(q_i, k_i)$
        \endtry
    \Else
        \State \textbf{prepare} $n_i$'s sub-question $q_i$, children nodes $c_i$
        \If {$c_i$ is $[DR]$}
            \State $a_i \gets SiblingAnswerReasoning(q_i, s_i)$
        \Else
            \State $a_i \gets ChildAnswerReasoning(q_i, c_i)$
            \If{$a_i$ is Unknown}
                \State $s_i \gets KnowledgeSourceSelection(q_i)$
                \State $k_i \gets KnowledgeRetrieval(q_i, s_i)$
                \State $a_i \gets DirectRAGReasoning(q_i, k_i)$
            \EndIf
        \EndIf
    \EndIf
\EndFor
\State \Return $a_{Root}$
\end{algorithmic}
\label{alg:reasoning_execution}
\end{algorithm}

\section{Distribution of Atomic Knowledge Operators}
\label{sec:distribution_operators}

\begin{figure}[!t]
  \noindent
  \centering
  \scalebox{1}{
  \includegraphics[width=\linewidth]{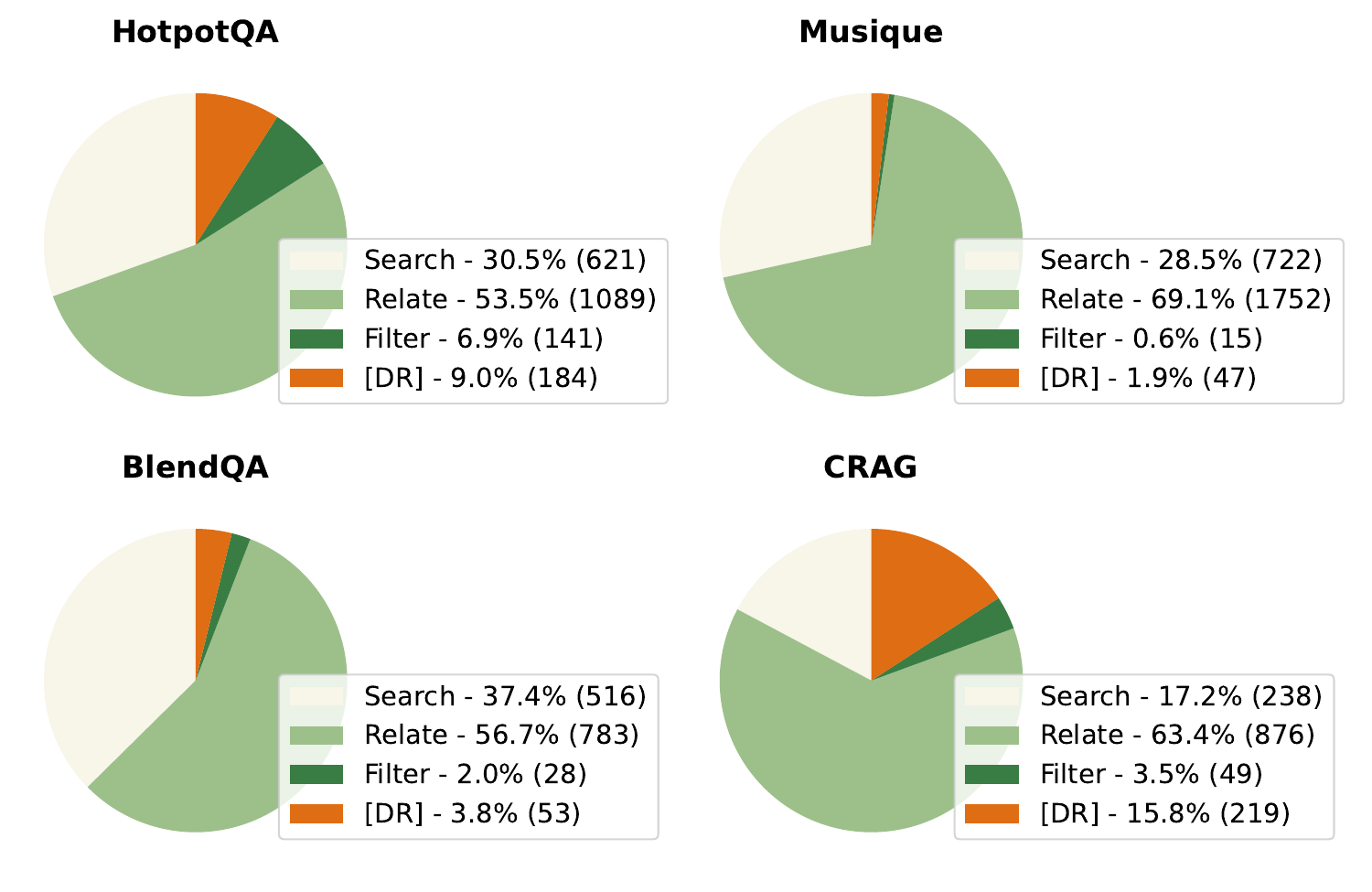}
  }
  \caption{Distribution of atomic knowledge operator calls.}
  \Description{}
  \label{fig:distribution_operators}
\end{figure}

Figure~\ref{fig:distribution_operators} illustrates the test-time distribution of \model's three atomic knowledge operators. 
\relate is the most frequently selected operator across all datasets, which is expected, as it serves for one-hop inference—the fundamental operation of multi-hop reasoning. 
In contrast, \filter is the least selected operator, with particularly low selection in Musique, where it is chosen only $0.6\%$ of the time. 
However, this is notable because Musique’s data construction process did not involve entity filtering operations. 
The minimal selection of \filter on Musique demonstrates \model's ability to intelligently adapt to differently constructed datasets by autonomously adjusting the use of the three operators.

\section{Distribution of Selected Knowledge Sources}

\label{sec:distribution_knowledge_sources}

\begin{figure}[ht]
  \noindent
  \centering
  \scalebox{1}{
  \includegraphics[width=\linewidth]{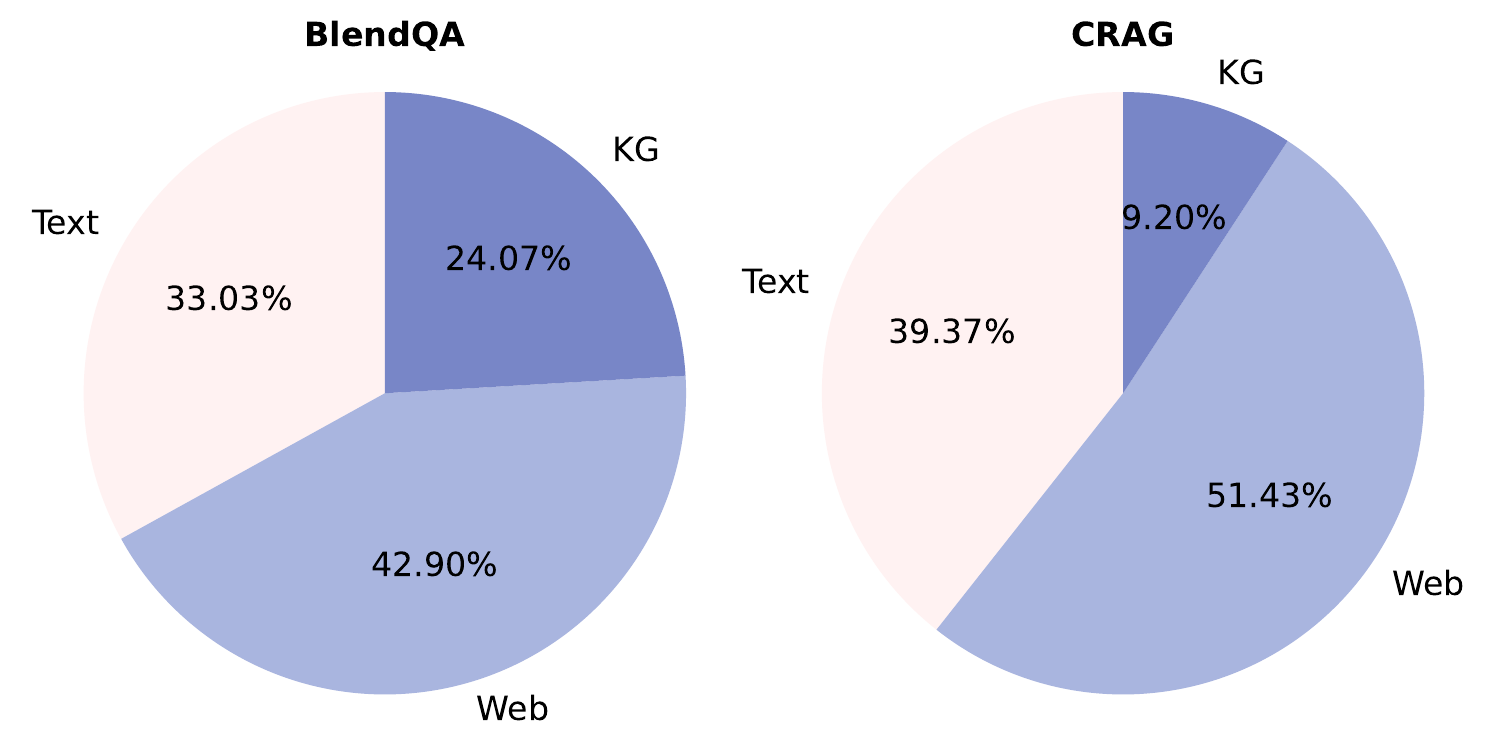}
  }
  \caption{Distribution of selected knowledge sources.}
  \Description{}
  \label{fig:distribution_knowledge_sources}
\end{figure}

Figure~\ref{fig:distribution_knowledge_sources} displays the test-time distribution of the three heterogeneous knowledge sources 
on the two muti-source datasets.
This distribution aligns highly to the actual structure of the datasets: BlendQA is constructed on balanced data from all three sources, while CRAG is constructed mainly on text and web sources. 
This demonstrates that AtomR yields a promising knowledge source selection strategy.

\section{Ablating backbone LLMs}
\label{sec:ablating_llms}

To investigate \model's performance across different LLM sizes, we conduct a backbone LLM ablation.
We randomly sample 100-entries from each dataset.
\begin{table}[htbp]
\caption{Ablating the backbone LLM}
\label{tab:ablating_llms}
\scalebox{0.95}{
\begin{tabular}{llll}
    \toprule
      & HotpotQA  & 2Wiki  & Musique \\
    \midrule 
    \multicolumn{3}{l}{\textit{GPT-4o}}  \\ 
    \midrule
    Standard Prompting &  50.77  & 45.30  & 21.04  \\  
    Standard RAG &  57.33  & 50.51  & 19.46  \\ 
    \textbf{\model} & 72.06\textcolor{darkgreen}{$(\uparrow14.7)$}  & 79.71\textcolor{darkgreen}{$(\uparrow29.2)$}  & 33.47\textcolor{darkgreen}{$(\uparrow26.9)$} \\
    \midrule 
    \multicolumn{3}{l}{\textit{Llama-3.1-70B-Instruct}}  \\ 
    \midrule
    Standard Prompting &  39.70  & 40.08  & 15.48 \\  
    Standard RAG &  49.43  & 42.41  & 14.42  \\ 
    \textbf{\model} & 56.72\textcolor{darkgreen}{$(\uparrow7.3)$}  & 72.15\textcolor{darkgreen}{$(\uparrow29.7)$}  & 29.92\textcolor{darkgreen}{$(\uparrow15.5)$} \\
    \midrule 
    \multicolumn{3}{l}{\textit{Llama-3.1-8B-Instruct}} \\
    \midrule 
    Standard Prompting &  27.41  & 28.25  & 6.39 \\  
    Standard RAG &  47.99  & 31.45  & 9.26  \\ 
    \textbf{\model} & 23.99\textcolor{darkred}{$(\downarrow24.0)$}  & 20.77\textcolor{darkred}{$(\downarrow10.7)$}  & 6.32\textcolor{darkred}{$(\downarrow2.9)$} \\
    \bottomrule
\end{tabular}
}
\end{table}

Table~\ref{tab:ablating_llms} illustrates the performance of \model across three backbone LLMs: GPT-4o, Llama-3.1-70B-Instruct, and Llama-3.1-8B-Instruct.
Results show that AtomR yields outstanding performance gains with Llama-3.1-70b-Instruct as the backbone, achieving significant F1 gains over the Standard RAG baseline on all three datasets.
However, AtomR leads to serious performance degradation with Llama-3.1-8b-Instruct. 
This shows that AtomR is a framework too complex for LLMs of small sizes. 
Yet, AtomR’s success with Llama-3.1-70b-Instruct demonstrates that it could yield promising results with 70b-level open-source models, which are widely accessible to the research community.

\section{Ablating text retriever models}
\label{sec:ablating_retrievers}

\model is a plug-and-play framework that works with any text retriever model.
We conduct an ablation on \model's text retriever model using the same 100-entry subsets from Appendix~\ref{sec:ablating_llms}.
Table~\ref{tab:ablating_retrievers} illustrates the performance of \model using three text retriever models: ColBERTv2,
BGE-large-en-v1.5, and BM25.
\begin{table}[!h]
\caption{Ablating the text retriever model}
\label{tab:ablating_retrievers}
\scalebox{0.95}{
\begin{tabular}{lcccc}
    \toprule
      & HotpotQA  & 2Wiki  & Musique & Avg. \\
    \midrule 
    \textbf{\model} (ColBERTv2)  &  \underline{72.06}  & \textbf{79.71}  & \textbf{33.47}  & \textbf{61.75} \\ 
    \midrule 
    \textbf{\model} (BGE-v1.5) & \textbf{74.42} & \underline{78.01}  & \underline{32.13}  & \underline{61.52} \\ 
    \textbf{\model} (BM25) & 66.97  & 76.81  & 29.99  & 57.92\\
    \bottomrule
\end{tabular}
}
\end{table}

Results show that \model yields comparative performance among the two dense retriever models ColBERTv2 and BGE-v1.5, but shows a slight performance degradation on the term-based retrieval model BM25.
Thus, it is best to implement \model with dense text retriever models to optimize its performance.

\section{Comparing \model to Additional Baselines}

Considering that \model is a RAG framework that supports (1) dynamic retrieval, and (2) KG-level atomic knowledge manipulation, we compare it against two additional baselines: (1) DRAGIN\cite{dragin}, a dynamic RAG framework with real-time retrieval decisions; (2) HippoRAG\cite{hipporag}, a RAG framework that combines LLM KG construction and KG-based RAG.
We experiment on the 100-entry subsets from Appendix~\ref{sec:ablating_llms} with Llama-3.1-70b-Instruct as the backbone.

\begin{table}[!h]
\caption{Comparing \model to additional baselines}
\label{tab:additional_baselines}
\scalebox{0.95}{
\begin{tabular}{lcccc}
    \toprule
      & HotpotQA  & 2Wiki  & Musique & Avg. \\
    \midrule 
    DRAGIN~\cite{dragin} & 23.4  & 56.6  & 19.3  & 33.1 \\ 
    HippoRAG~\cite{hipporag}  & 18.7  & 14.4  & \textbf{30.3}  & 21.1 \\ 
    \midrule
    \textbf{\model}  & \textbf{56.7}  & \textbf{72.2}  & 29.9  & \textbf{52.9} \\
    \bottomrule
\end{tabular}
}
\end{table}

Table~\ref{tab:additional_baselines} reports the F1 scores of DRAGIN, HippoRAG, and \model. 
Overall, \model significantly outperforms both methods, achieving respective F1 gains of 31.8\% and 19.8\%. 
Furthermore, on the HotpotQA subset, \model outperforms HippoRAG by achieving 303.2\% F1 performance with only 2.5\% completion token consumption.
This is because \model achieves KG-level fine-grained knowledge retrieval and reasoning without HippoRAG's expensive construction of an actual KG.
These results highlight the effectiveness and cost-efficiency of \model’s framework and operator design.

\section{\model Tree Depth Analysis}

To better understand the computational complexity of \model, we conduct a tree depth analysis of \model's Atomic Reasoning Trees (ARTs).
As shown in Table~\ref{tab:tree_depth_analysis}, \model produces ARTs with controllable complexity, with a maximum depth of 5 on HotpotQA and an average depth below 3 across all datasets.
Notably, the CRAG dataset exhibits the shallowest tree depth distribution, reflecting the simplicity of its questions compared to the other datasets.
These results indicate that AtomR is capable of adapting its tree depth dynamically to match the complexity of different questions.

\begin{table}[!h]
\caption{\model tree depth analysis.}
\label{tab:tree_depth_analysis}
\scalebox{0.95}{
\begin{tabular}{lcccc}
    \toprule
    Dataset  & Mean  & Median  & Min  & Max \\
    \midrule 
    HotpotQA & 2.65  & 3.00  & 2.00  & 5.00 \\ 
    2Wiki  & 2.97  & 3.00  & 2.00  & 3.00 \\ 
    Musique & 2.47 & 2.00  & 2.00  & 4.00 \\
    BlendQA & 2.42  & 2.00  & 2.00  & 4.00 \\
    CRAG & 1.84  & 2.00  & 1.00  & 3.00 \\
    \bottomrule
\end{tabular}
}
\end{table}

\begin{figure}[!b]
  \noindent
  \centering
  \includegraphics[width=\linewidth]{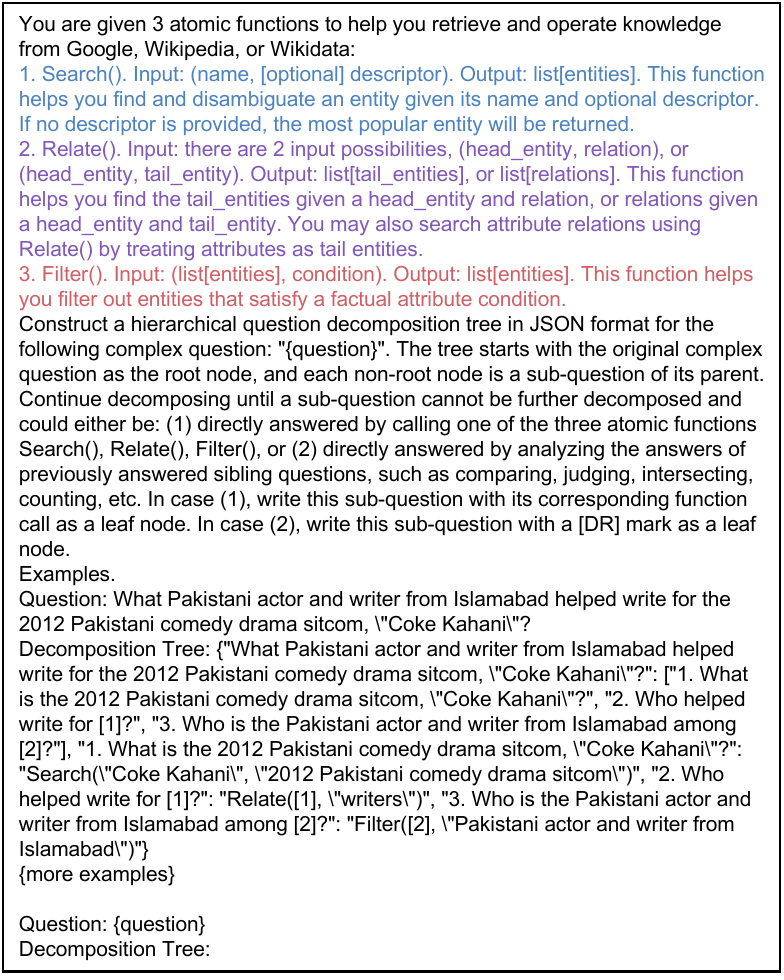}
  \caption{Prompt for Atomic Reasoning Tree generation.
  \label{fig:prompt_ART}}
  \Description{}
\end{figure}

\section{In-context Learning Prompts}
\label{sec:prompts}

\begin{figure}[!b]
  \noindent
  \centering
  \includegraphics[width=\linewidth]{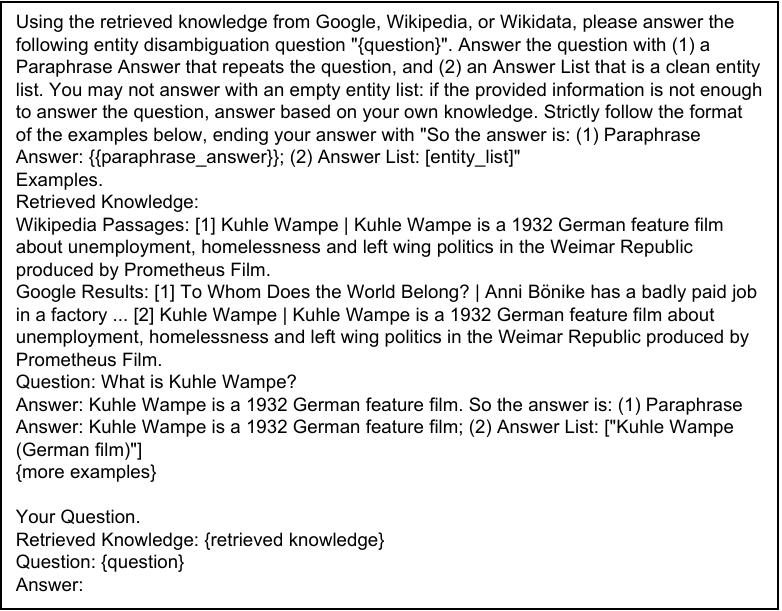}
  \caption{Prompt for the adaptive LLM executor of the \search operator.}
  \label{fig:prompt_search}
  \Description{}
\end{figure}

\begin{figure}[!b]
  \noindent
  \centering
  \includegraphics[width=\linewidth]{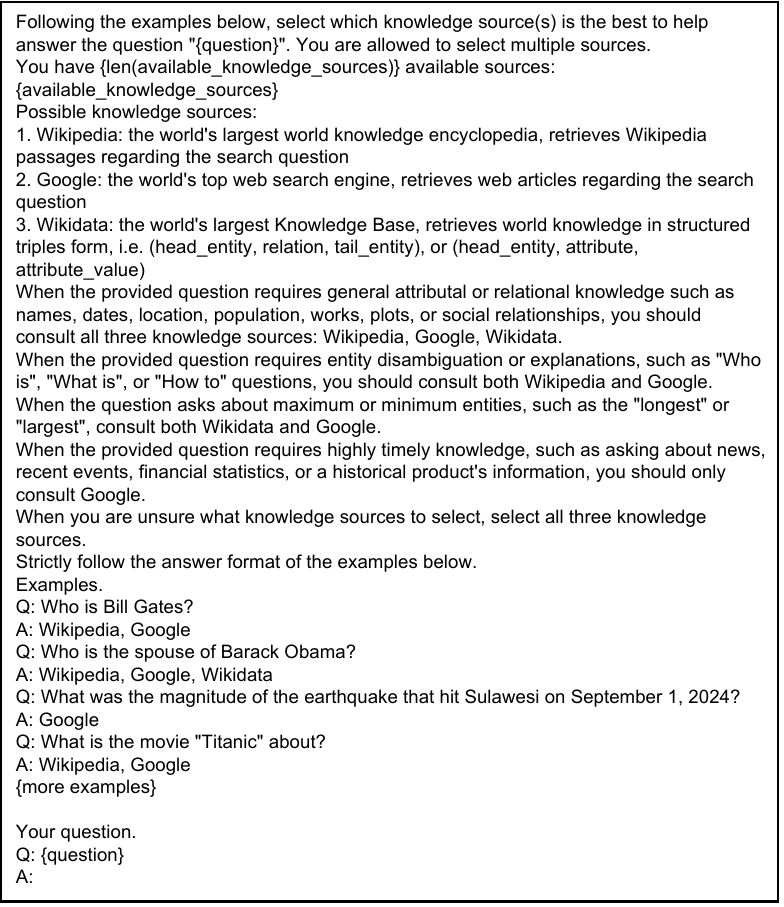}
  \caption{Prompt for Dynamic Knowledge Source Selection.}
  \Description{}
  \label{fig:prompt_knowledge_selection}
\end{figure}

Figures~\ref{fig:prompt_ART} to~\ref{fig:prompt_knowledge_selection} display representative
LLM prompts of \model.
For all prompts, please refer to our official code repository.

\end{document}